\documentclass[final,5p,times,twocolumn]{elsarticle}
\usepackage{amssymb}
\usepackage{amsmath}
\usepackage{balance}
\usepackage{multirow}
\usepackage{booktabs}
\usepackage{makecell}
\newcommand{\Rmnum}[1]{\uppercase\expandafter{\romannumeral #1}} 

\begin{document}

\begin{frontmatter}




\title{Joint Moment Estimation for Hip Exoskeleton Control: 
\\ A Generalized Moment Feature Generation Method}

\author[1]{Yuanwen Zhang \fnref{fn1}}
\author[1]{Jingfeng Xiong \fnref{fn1}}
\author[1]{Haolan Xian}
\author[1]{Chuheng Chen}
\author[2]{Xinxing Chen}
\author[1]{Chenglong Fu}
\author[1]{Yuquan Leng\corref{cor1}}

\affiliation[1]{organization={Shenzhen Key Laboratory of Biomimetic Robotics and Intelligent Systems and Guangdong Provincial Key Laboratory of Human-Augmentation and Rehabilitation Robotics, Department of Mechanical and Energy Engineering, Southern University of Science and Technology},
city={Shenzhen},
country={China}}
\affiliation[2]{organization={The Key Laboratory of Image Processing and Intelligent Control and the Hubei Key Laboratory of Brain-inspired Intelligent Systems, School of Artificial Intelligence and Automation, Huazhong University of Science and Technology},
city={Wuhan},
country={China}}
\cortext[cor1]{Corresponding author (e-mail: yuquanleng@163.com)}
\fntext[fn1]{These authors contributed equally to this work.}
\begin{abstract}
Hip joint moments during walking are the key foundation for hip exoskeleton assistance control. Most recent studies have shown estimating hip joint moments instantaneously offers a lot of advantages compared to generating assistive torque profiles based on gait estimation, such as simple sensor requirements and adaptability to variable walking speeds. However, existing joint moment estimation methods still suffer from a lack of personalization, leading to estimation accuracy degradation for new users. To address the challenges, this paper proposes a hip joint moment estimation method based on generalized moment features (GMF). A GMF generator is constructed to learn GMF of the joint moment which is invariant to individual variations while remaining decodable into joint moments through a dedicated decoder. Utilizing this well-featured representation, a GRU-based neural network is used to predict GMF with joint kinematics data, which can easily be acquired by hip exoskeleton encoders. The proposed estimation method achieves a root mean square error of 0.1180 ± 0.0021 Nm/kg under 28 walking speed conditions on a treadmill dataset, improved by 6.5\% compared to the model without body parameter fusion, and by 8.3\% for the conventional fusion model with body parameter. Furthermore, the proposed method was employed on a hip exoskeleton with only encoder sensors and achieved an average 20.5\% metabolic reduction ($p<0.01$) for users compared to assist-off condition in level-ground walking.
\end{abstract}




\end{frontmatter}

\section{Introduction}
Lower-limb exoskeletons are capable of assisting in human locomotion, thus reducing joint moments, metabolic expenditure, and mechanical loading of joints. This allows it to be applied to assist weight-bearing workers, elderly individuals in need of walking assistance, and patients with arthritis \cite{caoLowerLimbExoskeleton2022, hyunBiomechanicalDesignAgile2017,kalitaDevelopmentActiveLower2021,kapsalyamovStateArtLower2019,medranoBiologicalJointLoading2021}. During walking, the hip joint contributes 40-50\% of the total positive joint power of the lower limb joints \cite{farrisMechanicsEnergeticsHuman2012}, which means that hip-assisted exoskeletons can bring substantial benefits. Furthermore, compared with other lower limb exoskeletons, hip exoskeletons have smaller metabolic penalties caused by distal-borne mass \cite{kimReducingMetabolicRate2019}. Therefore, hip exoskeletons are considered an important choice for walking assistance.

The moment of the hip joint is a fundamental basis for hip exoskeleton assistance control. Since calculating joint moments with motion capture systems or force platforms is not feasible for the hip exoskeleton in the real world, it is necessary to utilize other sensors fixed on the exoskeleton system. Recent researchers have focused on determining the assistive torque of the exoskeleton based on the human motion information measured by these sensors, such as encoders on the motors or inertial measurement units (IMUs).

Currently, there are two main methods to generate the assistive torque of hip exoskeletons based on hip joint motion data: determining output torque based on assistive torque curve generation and gait phase estimation, and calculating assistive torque according to instantaneous joint moment estimates. 
The former method involves the following steps: (a) estimating the gait phase and the assistive torque profile for the entire gait cycle based on the user's historical walking data, and (b) generating the assistive torque based on the gait phase and the assistive torque profile~\cite{xueAdaptiveOscillatorBasedRobust2019,qianAdaptiveOscillatorBasedAssistive2022,seoAdaptiveOscillatorBasedControl2018a, molinaroEstimatingHumanJoint2024}. However, this method may exhibit delayed assistive torque when dealing with variable walking gaits, which can hinder the user's movements. The other method generates assistive torque directly through a computational model by measuring the kinematic information of the human body \cite{limDelayedOutputFeedback2019,zhangExperimentalComparisonTorque2015,gasparriProportionalJointMomentControl2019}. Compared to the first method, the second method calculates assistive torque independent of gait periodicity, allowing it to generate appropriate walking assistance in real-time even with changing gaits. Therefore, calculating assistive torque with instantaneous joint moment estimation has the potential to become an important computational method for future exoskeleton control.

Estimatign the hip biomechanical joint moment accurately is the key foundation to calculate assistive torque instantaneously, but it remains a significant challenge to obtain the estimates based on measurable data from the wearable sensors on the human body. Although some researchers have calculated joint moments by establishing approximate dynamic models \cite{buchananEstimationMuscleForces2005, bisheAdaptiveAnkleExoskeleton2021}, these methods rely on complex sensor equipment such as ground reaction force sensors, which increases the system complexity, leading to the instability of the exoskeleton system. Besides, extra sensors can be uncomfortable for the users to wear and even hinder their movement.

As data-driven methods have shown remarkable effectiveness in biological measurement and exoskeleton control fields \cite{joshiMoveNetDeepNeural2021, mundtPredictionLowerLimb2020, qianPredictiveLocomotionMode2022, zhangBoostingPersonalisedMusculoskeletal2022}, researchers have started focusing on measured motion information (such as hip joint position changes) by simple and lightweight sensors during human walking for the estimation of joint moment. Wang used the XGBoost method to achieve the estimation of the knee joint adduction moment in an experiment involving 106 participants by using only the information from two IMU sensors \cite{wangRealTimeEstimationKnee2020}. However, traditional machine learning methods have limited performance in improving the estimation accuracy of joint moment estimation, and researchers are turning to neural networks to further enhance the estimation accuracy of joint moment. Molinaro trained a TCN network that utilized expanded causal convolutional layers to encode temporal information and learn data features from input, achieving hip joint moment estimation based on data of simulated IMU and angle of hip joint \cite{molinaroSubjectIndependentBiologicalHip2022, molinaroEstimatingHumanJoint2024}. Furthermore, Hossain designed an Ensemble network that utilized deep neural network layers with different structure types, improving the performance of human dynamics estimation. Through fusion modules and bagging techniques, the model achieved a more accurate estimation of lower-limb joint moment \cite{hossainEstimationLowerExtremity2023}.


For the hip exoskeleton control, the joint moment estimation model should be able to adapt to new users so it can provide accurate joint moment estimation in real-time. However, the existing methods have not yet achieved this goal, for the reasons that the influence of human parameters is underestimated or difficult to obtain an estimation model that integrates individual characteristics.
The lack of personalization might lead to poor applicability to new users, where the assistance torque generated by the exoskeleton might not be suitable for the user's walking. 

To address the above problems, we propose a joint moment estimation neural network based on generalized moment feature (GMF) generation. A feature generator was designed to obtain GMF, which is more generalized than the joint moment, facilitating the estimator to learn the relationship between the kinematics and dynamics during human walking better. Meanwhile, the GMF can be decoded with body parameters into the joint moment, achieving personalizing joint moment estimation. Furthermore, a GRU-based estimator was designed to reduce model computation time with minor accuracy loss under the proposed GMF neural network framework.

The main contributions of this study are as follows:
\begin{itemize}
\item[1)]
Proposing a novel joint moment estimation neural network based on user-specific body parameters, which is conducive to obtaining a superior individual-independent representation of biomechanical joint moments and achieving better estimation accuracy.
\end{itemize}

\begin{itemize}
\item[2)]
Developing the proposed method on a hip exoskeleton to validate its assistance gain for users as well as showing its potential application value in scenarios relying solely on encoder sensors.
\end{itemize}


The structure of this paper will be described as follows. Section \Rmnum{2} describes the proposed joint moment estimation method based on the GMF generation. The experiments and results are presented in sections \Rmnum{3} and \Rmnum{4}, discussed in section \Rmnum{5}. The conclusion of this paper is written in section \Rmnum{6}.

\section{Method}

\begin{figure*}[h]
  \centering 
  \includegraphics[width=\linewidth]{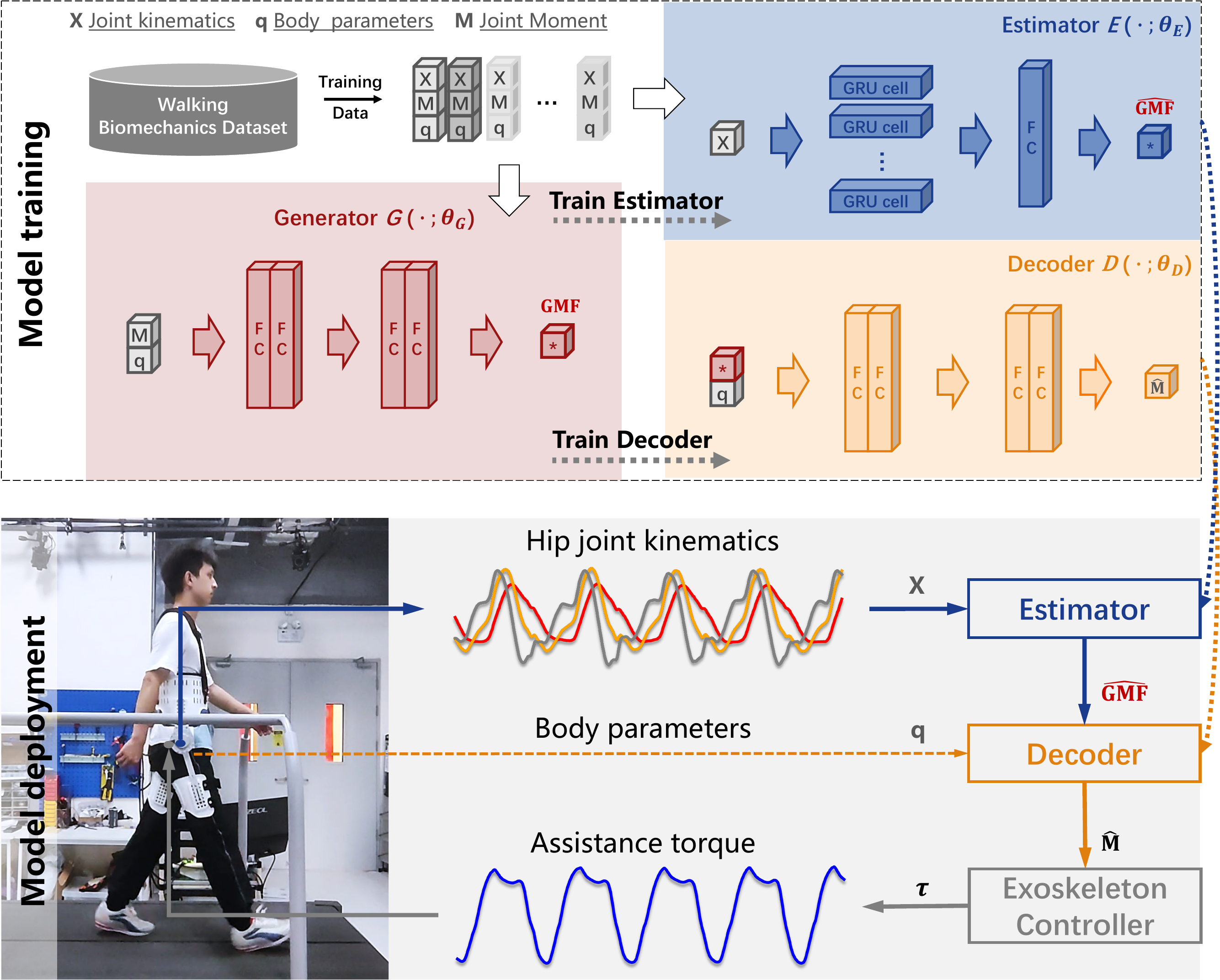}  
  \caption{Joint moment estimation for hip exoskeleton assistance based on GMF generation. During the training process, kinematics information ($\boldsymbol{X}$) is utilized as input for the estimator, while body parameters ($\boldsymbol{q}$) and hip joint moments ($M$) serve as inputs for the generator. The decoder takes body parameters and the GMF generated by the generator (represented by "*") as inputs. The estimator and decoder are then employed to estimate hip joint moments ($\hat{M}$) to provide assistance torque during walking for new users (the generator is not needed).}
  \label{overview}
\end{figure*}

This section provides a detailed explanation of the proposed joint moment estimation method based on GMF generation. We will first describe the conventional form of using neural networks to estimate biomechanical joint moments. Then we introduce the concept of GMF, addressing the issues that arise from the conventional form. Based on this, we propose a neural network framework for generating GMF and estimating it using a GRU module, as well as a decoder for decoding it into the joint moment (Fig.\ref{overview} - Model training). Finally, we present the training strategy for the proposed neural network model and the deployment of estimating joint moment in the real world (Fig.\ref{overview} - Model deployment).
\subsection{The Basic Model of Joint Moment Estimation}
This study aims to estimate the sagittal hip joint moments based on the joint kinematics during walking utilizing neural networks. As mentioned in \cite{zhangSurveyNeuralNetwork2021}, the interpretability of a model is crucial for generating models with superior accuracy and robustness. Therefore, two essential attributes are assumed that a neural network model should possess to estimate joint moment: (a) incorporating the relationship between kinematics and dynamics, and (b) ensuring the continuity of motion.

The rationality of the second attribute is supported by previous researches \cite{eslamyMultiJointLegMoment2023, molinaroAnticipationDelayedEstimation2023,zhang2023predict}, which indicates that using an appropriate window to capture the input time series can improve the estimation accuracy of the joint estimation model. Regarding the first attribute, to determine the model inputs from candidate kinematic information, we conducted a series of preliminary experiments. Data comprising the angles, angular velocities, angular accelerations, and their combinations of the bilateral hip joints were selected as the inputs for each time step of the estimator. The results of these experiments demonstrated that as the dimensionality of the inputs increased, the accuracy of the joint moment estimation model gradually improved. Therefore, the general inputs of the joint moment estimation model can be described as follows: $\boldsymbol{X}=[\boldsymbol{x}_t,\boldsymbol{x}_{t-1},...,\boldsymbol{x}_{t- \Delta T} ] \in R^{ \Delta T \times D}$, where $\Delta T$ represents the window size and $D$ denotes the dimensionality of the joint kinematics data ($\Delta T=100$, $D=6$). A basic neural network model for estimating joint moments is described as $F$. For the estimated joint moment $\widehat{M}$ at time $t$, the process of estimation can be expressed as:
\begin{equation}
  \widehat{M} = F(\boldsymbol{X}).
  \label{eq_M_Fx}
\end{equation}

\subsection{GMF-based Model}

The conventional form of joint moment estimation networks does not incorporate personalized information of the subjects as the individual parameters are not included in (\ref{eq_M_Fx}). Therefore, to address the issue of decreased accuracy when estimating joint moments using only joint kinematics in a multi-user environment, we propose an novel joint moement estimation method based on GMF generation. 

\subsubsection{Inspiration of GMF Generation}
Firstly, we introduce the concept of GMF, which is an individual-independent representation from the joint moment that combines with the subject's body parameters $\boldsymbol{q}$ (mass $m$ and height $h$ are used in this paper, which means $\boldsymbol{q} = [m, h]$). Similar to the expression in (\ref{eq_M_Fx}), the joint moment estimation method based on GMF can be represented as follows:
\begin{equation}
    GMF = G(\boldsymbol{q}, M)
  \label{eq_gmf_g}
\end{equation}
\begin{equation}
  \widehat{GMF} = E(\boldsymbol{X})
  \label{eq_gmf_e}
\end{equation}
\begin{equation}
  \widehat{M} = D(\widehat{GMF} ),
  \label{eq_gmf_d}
\end{equation}

\noindent where $M$ denotes the ground truth of joint moment, $G$ denotes the GMF generator, $E$ denotes the GMF estimator and $D$ denotes the GMF decoder, satisfying $D (E (\cdot)) \in F (\cdot)$. 

Equations of (\ref{eq_gmf_g}), (\ref{eq_gmf_e}), and (\ref{eq_gmf_d}) show the core framework of GMF-based model to achieve a better estimation: estimating the GMF instead of directly estimating the joint moment, so that the decrease in estimation accuracy caused by individual differences can be reduced. In other words, GMF-based model improves the generalization performance of the joint moment estimation across multiple users.

Based on the these inspiration, we design a method for generating GMF and apply it to the training of the joint moment estimation model. Fig.~\ref{overview} illustrates the process of generating GMF. The joint moment estimation method based on GMF leverages both the joint moment and body parameters. It utilizes an adaptive neural network to extract GMF that is independent of individuals. This feature is then used to train the estimator. Additionally, a decoder is trained to decode the estimated GMF into the joint moment. It is worth noting that, compared with inputting body parameters into the network directly, this method expresses the role of body parameters in the network model more specifically, which means it can achieve better optimization from training.

\subsubsection{GMF Generator}
A feature generator $G$ is used to generate the GMF. More specifically, this generator utilizes a neural network to combine the joint moment of subjects with their body parameters and finally generates the desired GMF.

The main objective of GMF generation is to discover a better estimation target to achieve a more accurate estimation by the estimator $E$ as described by \eqref{eq_gmf_e}. Therefore, unlike typical deep learning tasks, in this method, the labels learned by the estimator and the GMF generator are mutually the output values of each other during the training process. This structure guides the GMF generator to generate training labels, namely the proposed GMF, from the joint moment and body parameters of subjects, which can enhance the accuracy of the estimator.

Meanwhile, it is essential that the GMF generated by the generator should be able to be decoded into the joint moment. Using $\boldsymbol{\theta}_{G}$ to represent the network parameters of the feature generator $G$, 
the optimal generator parameters $\boldsymbol{\theta}_{G}^*$ in the above process can be mathematically represented as:
\begin{equation}
\begin{aligned}
  \boldsymbol{\theta}_{G}^* = \mathop{\textbf{argmin}}\limits_{\boldsymbol{\theta}_{G}} &\left \| E(\boldsymbol{X}) - G(\boldsymbol{q},M;\boldsymbol{\theta}_{G})\right \|_2 \\
  &s.t. ~G(\boldsymbol{\theta}_{G}) \in \boldsymbol{\mathcal{G}},
\end{aligned}
\label{eq_g_argmin}
\end{equation}
where $\boldsymbol{\mathcal{G}}$ represents the function space to which $G$ belongs, ensuring the GMF generated through $G$ can be decoded into joint moment.

The network structure of the training phase (as shown in Fig. \ref{overview}) provides a specific illustration of the role of the GMF generator within the overall network structure. The inputs of the generator include joint moment and body parameters, and its output is the GMF. The GMF Generator serves two important functions: (a) generating the optimal GMF to improve the estimation accuracy of the estimator, enhancing the model's performance across different users; (b) ensuring the decoder can accurately decode the generated GMF into the joint moment.
To implement a model with the proposed method, as an example, we employ a fully connected network with five hidden layers to construct the GMF generator. Each hidden layer consists of 32 nodes, and a LeakyReLU activation function is applied after each hidden layer to achieve non-linearity.

\subsubsection{GMF Estimator}

In exoskeleton systems, recurrent neural networks (RNN) can better leverage their rapid computation ability~\cite{yamak2019comparison} to estimate joint moments.
Existing recurrent neural network structures include RNN, LSTM, GRU \cite{choPropertiesNeuralMachine2014}. Due to the challenges of RNN in handling long-term dependencies, vanishing gradients, and exploding gradients, they have been gradually replaced by LSTM and GRU. Furthermore, GRU has a more simplified structure and fewer parameters compared to LSTM, resulting in lower memory usage and faster computation efficiency. In many cases, there is no significant difference in estimation accuracy between the two models \cite{chungEmpiricalEvaluationGated2014}. Therefore, the estimator in this study utilizes a neural network based on the GRU structure.

\begin{figure}[h]
  \centering 
  \includegraphics[width=3.48in]{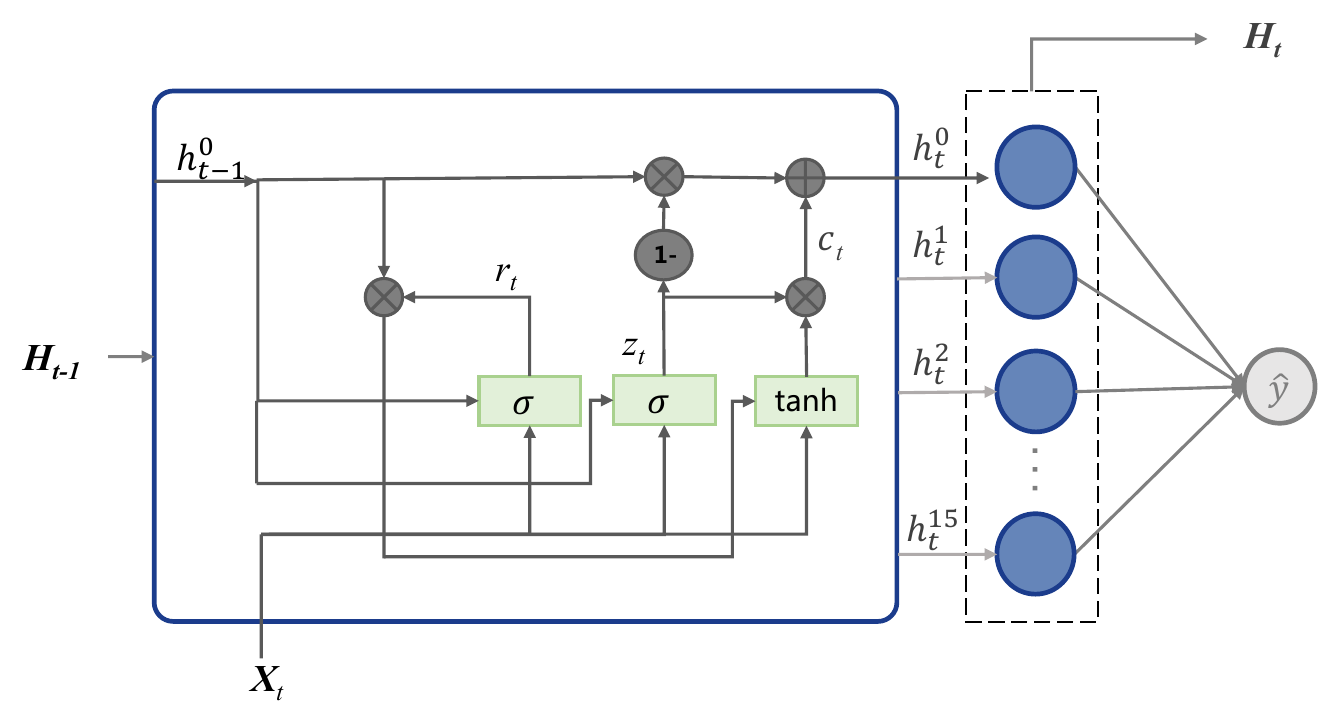}  
  \caption{The fast estimator based on GRU, the computation process of one output within the GRU unit, denoted as $h^0$, is shown in the solid box. The computation process for the remaining 15 hidden layer nodes, inputs ($h_{t-1}^1$ - $h_{t-1}^{15}$), and outputs ($h_t^1$ - $h_t^{15}$), follows the same process.}
  \label{estimator}
\end{figure}

In this study, only one GRU block is used as an implementation case, with 16 hidden neurons, and a fully connected network with a size of 16 $\times$ 1 is used to output the predicted target (Fig. \ref{estimator}). Specifically, at each time step, the GRU receives two different modalities: the hidden information from the previous time step's GRU hidden layer and the inputs $\boldsymbol{X}$. It adjusts the information flow through update gates and reset gates, and finally obtains the output based on these two gates. Thus, given the hip joint sagittal plane angle signal at time t as $\boldsymbol{X}_t$, the hidden inputs from the previous time step as $\boldsymbol{H}_{t-1}$, the estimated GMF at the current time step $\widehat{GMF}$ can be obtained as follows:
\begin{equation}
  \boldsymbol{H}_t = \text{GUR}(\boldsymbol{X}_t, \boldsymbol{H}_{t-1}),
\end{equation}
\begin{equation}
  \widehat{GMF} =\text{FC}(\boldsymbol{H}_t).
\end{equation}
Writing the network parameters ($\boldsymbol{W}$ and $\boldsymbol{b}$) as $\boldsymbol{\theta}_{E}$, like (5), the optimal parameters $\boldsymbol{\theta}_{E}^*$ are solved as:
\begin{equation}
  \boldsymbol{\theta}_{E}^* = \mathop{\textbf{argmin}}\limits_{\boldsymbol{\theta}_{E}} \left \| E(\boldsymbol{X};\boldsymbol{\theta}_{E}) - G(\boldsymbol{q},M;\boldsymbol{\theta}_{G})\right \|_2.
  \label{eq_e_argmin}
\end{equation}

\subsubsection{GMF Decoder}

The GMF decoder is used to decode the generated GMF into joint moment. We employ the same network structure as the GMF generator for constructing the GMF decoder. To improve the accuracy of the output, the body parameters are included as part of its inputs. Using $\boldsymbol{\theta}_{D}$ to represent the network parameters of the GMF decoder, 
the optimal parameters $\boldsymbol{\theta}_{D}^*$ of the GMF decoder in the above process can be obtained from:
\begin{equation}
  \boldsymbol{\theta}_{D}^* = \mathop{\textbf{argmin}}\limits_{\boldsymbol{\theta}_{D}} \left \| D(\boldsymbol{q},GMF;\boldsymbol{\theta}_{D}) - M\right \|_2.
  \label{eq_d_argmin}
\end{equation}

\subsection{Model Training}
The proposed method consists of two parts: GMF estimation and GMF decoding, which involves the training of three models: estimator, generator, and decoder. In the first part, the GMF generator is jointly trained with the estimator to obtain better individual-independent representation (GMF) and its estimation model (estimator). Meanwhile, in the second part, the GMF generator is trained together with the decoder to ensure the decodability of the generalized GMF and to obtain a model capable of decoding the GMF into the joint moment.

Based on (\ref{eq_g_argmin}), (\ref{eq_e_argmin}), and (\ref{eq_d_argmin}), two loss functions $L_1$, $L_2$ are constructed for the network training. The backpropagation process is shown in Fig.~\ref{overview} - Model training with solid lines. $L_1$ represents the loss between the results of the GMF estimator and the GMF generator, while $L_2$ represents the loss between the outputs of the GMF decoder and ground-truth values of the joint moment. The loss functions are defined as follows:
\begin{equation}
  J_{mse}(y_i, \hat{y_i}) = \frac{1}{N}\sum_{i=1}^{N} (y_i - \hat{y_i})^2,
\end{equation}
\begin{equation}
  L_1 = J_{mse}(F(\boldsymbol{X}), G(\boldsymbol{q},M)),
\end{equation}
\begin{equation}
  L_2 = J_{mse}(D(G(\boldsymbol{q},M),\boldsymbol{q}), M).
\end{equation}

It is worth noting that the optimization of the parameters of the GMF generator is also included in $L_2$. By selecting the weighting coefficients $w_1$ and $w_2$ of the two loss functions, it can be ensured that the GMF, which is generated by the feature generator and optimized by $L_1$, can be decoded into the joint moment through the GMF decoder. Therefore, the condition in (\ref{eq_g_argmin}) is satisfied. The overall loss function of the joint moment estimation network is given by:

\begin{equation}
  Loss = w_1 * L_1 + w_2 * L_2.
  \label{eq_total_loss}
\end{equation}

\subsection{Model Deployment}
The trained GMF estimator and decoder will be used in the estimation process for the final joint moment estimation. As shown in Fig. \ref{overview} - Model deployment, using the joint kinematics as input, the GMF is obtained by the fast GMF estimator, which is then combined with user body parameters to be decoded into the predicted joint moment:
\begin{equation}
  \widehat{M} = D(E(\boldsymbol{X}), \boldsymbol{q}).
  \label{eq_model_deploy}
\end{equation}

Subsequently, the estimated joint moments are converted into assistive torques by the mid-level and low-level controllers. In this study, we implement a three-step middle-level exoskeleton controller \cite{molinaroEstimatingHumanJoint2024,kang2019effect} to provide users assistance during walking.

\begin{figure}[t]
  \centering 
  \includegraphics[width=3.48in]{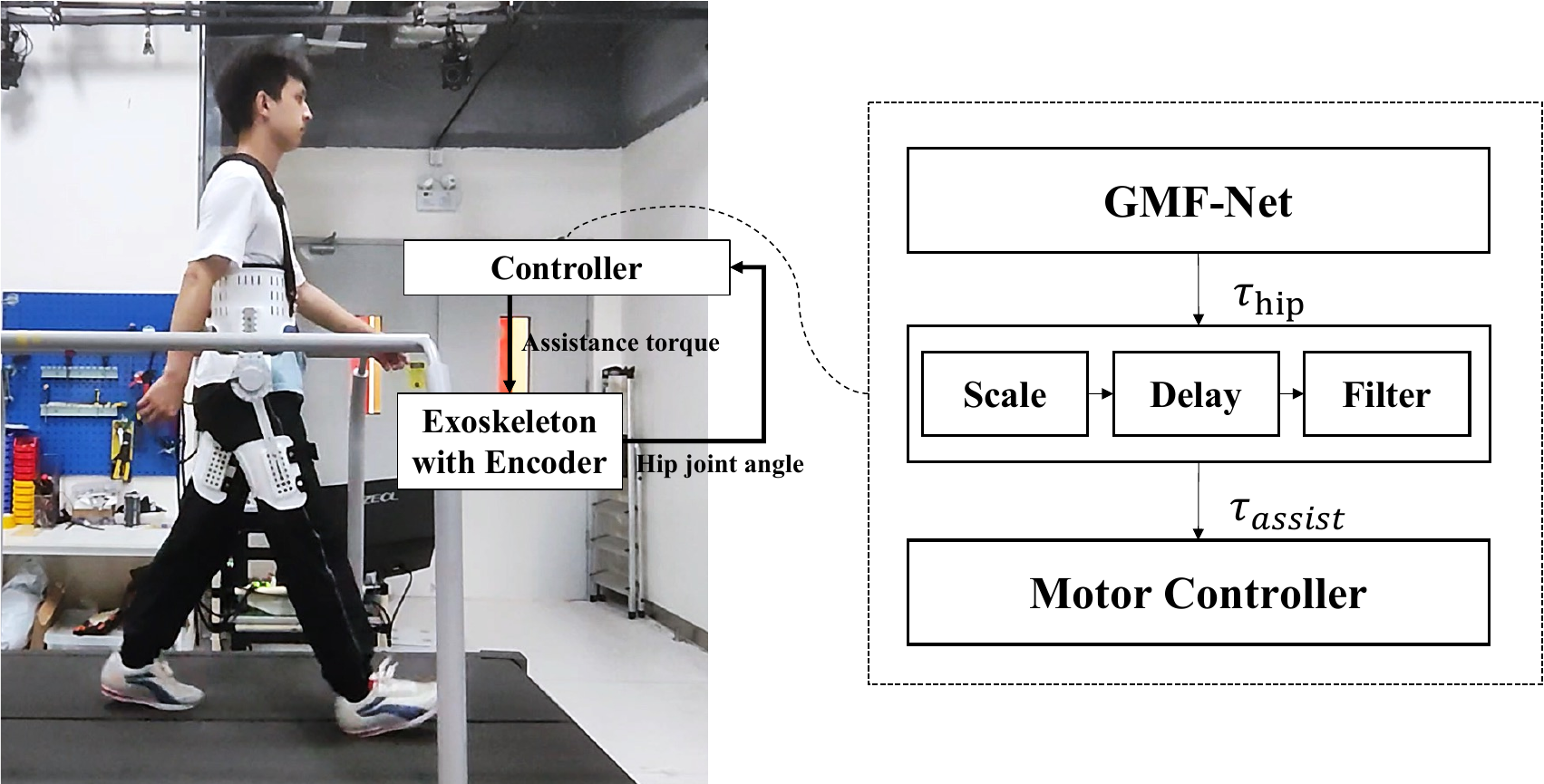}  
  \caption{Exoskeleton and controller used for validating the proposed method. Motors with encoder provide the controller hip joint angles as inputs and generate assistance torque to assist users. The controller is a three-level structure, where the high-level estimates hip joint moments of the user, the mid-level processes hip joint moments to obtain the desired assistance torque and the low-level FOC generates motor torque stably.}
  \label{exoskeletonController}
\end{figure}

A hip exoskeleton prototype was designed to implement assistance experience for users. As shown in Fig.~\ref{exoskeletonController}, the assistance torque is obtained by scaling, delaying, and filtering the hip moment estimates. Specifically, $\tau_\text{assist}$ at time $t$ is calculated by:
\begin{equation}
    \tau_{\text{assist},t} = \text{Low-pass Filter}(k \cdot \widehat{M}_{t-\delta}), 
    \label{eq_assist_tau}
\end{equation}
where $k$ is the scale factor and $\delta$ is the delay factor.

\section{Experiments}
There were two designed experiment parts in this study: the first part was to validate the GMF-based model by a dataset, which is aimed at evaluating the accuracy and generalization of the proposed method, and the second part was to apply the trained model on a hip exoskeleton to validate its performance (assistance gain for users) when used as an exoskeleton high-level controller. 

This work involved human subjects in its research. Approval of all ethical
and experiment procedures and protocols was granted by Medical Ethics
Committee of Southern University of Science and Technology [No.20240022], under Application 2024-02-24.
All participants provided written informed consent to participate in the study. 
\subsection{Implementation Details}
We conducted model training on a device with a CPU (Intel, Core i7-12700F) and an GPU (Nvidia, GeForce RTX 3060Ti). The Adam optimizer \cite{kingma2014adam} was employed. In all experiments, we performed five repetitions with different random initializations to ensure robustness. Detailed parameters are described in each section before the corresponding experiment results. The model deployment (exoskeleton controller) was implemented on a Mini-PC (Intel, NUC11PH).

\subsection{Validation on dataset}
\subsubsection{Dataset and Training-testing Data Partitioning}
The publicly available dataset used in this article is sourced from \cite{camargoComprehensiveOpensourceDataset2021}. The treadmill data of 20 healthy subjects (12 males and 8 females, age 21.20 ± 3.04 yr, height 1.70 ± 0.07 m, weight 67.76 ± 11.64 kg) were used for model training and testing. Specifically, the data of each subject included hip joint sagittal plane angle positions and flexion/extension joint moment at 28 different treadmill speeds. Data collection of each subject was performed through 7 experiments, with each experiment consisting of four different speeds maintained for 30 seconds each. The speeds for the first experiment were 0.5 m/s, 1.2 m/s, 1.55 m/s, and 0.85 m/s. In subsequent experiments, each speed increased by increments of 0.05 m/s. Hip joint angular velocity and angular acceleration were computed corresponding to the data collection frequency, and the hip joint moments used were smoothed using butterworth filtering.

\begin{figure}[!]
  \centering 
  \includegraphics[width=\linewidth]{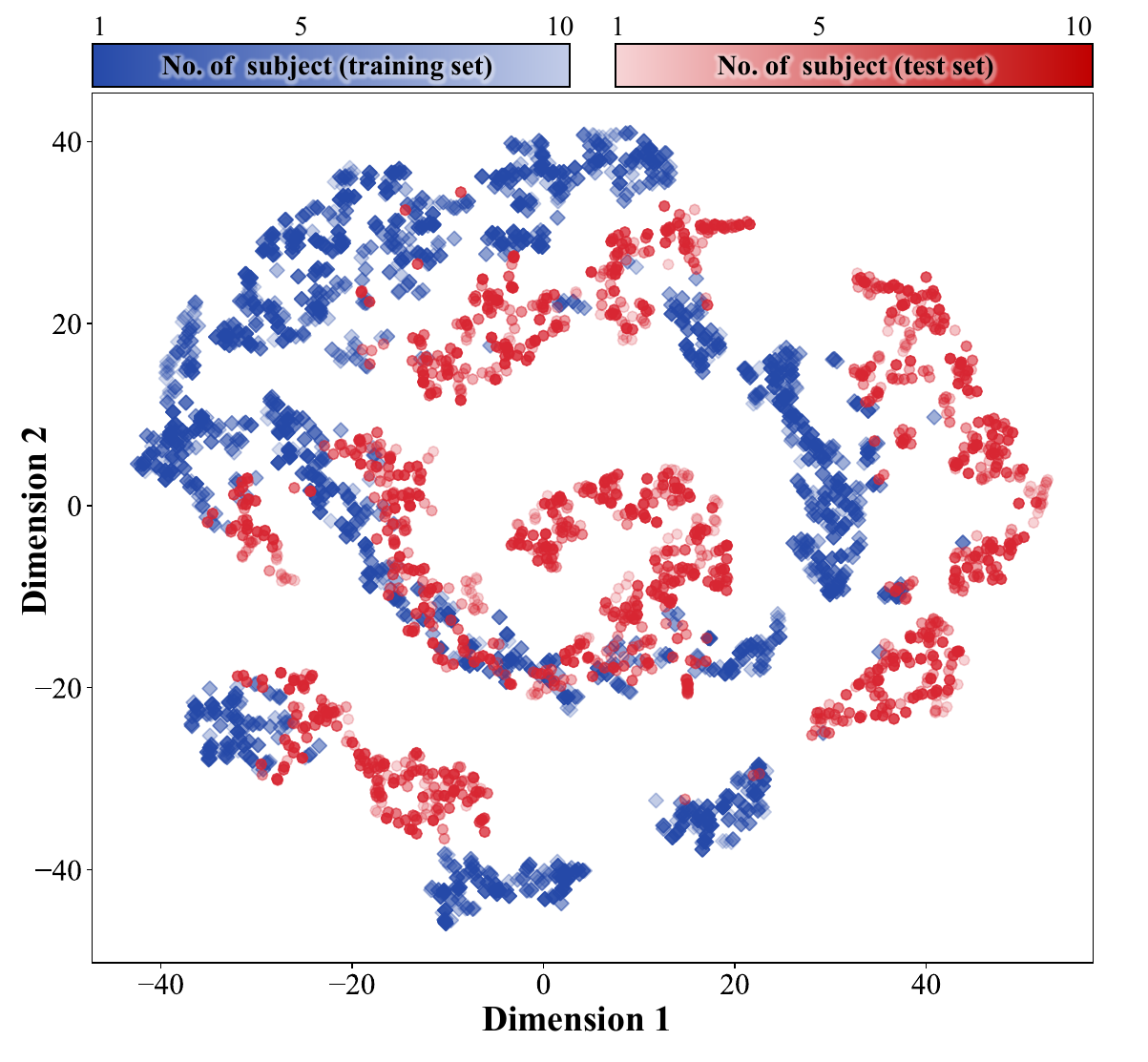}  
  \caption{Visualization of the input features distribution (joint kinematics) of the training set (in blue) and the test set data (in red) using the T-SNE method. The varying shades of color represent data from 10 individual subjects in each set. Under the data partitioning method that maximizes individual differences, the intra-subject distribution of input features within the same set (intra-class distance) appears to be similar, while the distribution of input features between different sets (inter-class distance) exhibits significant differences.}
  \label{tsne}
\end{figure}

To demonstrate the effectiveness of the proposed method, the dataset was partitioned based on inter-subject differences to create a maximum separation between training and test data, resembling a multi-user scenario. The T-SNE method was utilized to divide the hip joint angle data samples from the subjects into two categories, as shown in Fig. \ref{tsne}, with 10 subjects in each category. Consequently, for the experiments in this study, the training and validation sets were composed of data from the first 10 subjects (in an 8:2 ratio, independent of each other), while the data from the remaining 10 subjects were used as the test set to assess the model's performance.

\subsubsection{Models for accuracy validation}
We evaluated the performance of the proposed method by computing the RMSE of the predicted joint moment on the test dataset. To better demonstrate the effectiveness of the proposed method, we compared our method with two baseline methods: 
\begin{itemize}
  \item \textbf{Model without body parameters} - means directly estimating moment by inputting joint kinematics into the estimator without body parameters.
  \item \textbf{Fusion model without GMF} - means implementing a conventional fusion model where two networks separately take joint kinematics and body parameters as inputs to obtain two hidden features, and these two features are then stacked and passed through a fully connected network to estimate joint moment. 
\end{itemize}

To ensure comparability, the architecture of the part inputting joint kinematics in two baseline neural networks is the same with the GMF estimator, and the optimizer (Adam) parameters and initial learning rate ($lr = 0.001$) are the same. In our method, we set the parameters $w_1 = 1$ and $w_2 = 0.05$ in (\ref{eq_total_loss}).
The baseline models are represented as:
\begin{equation}
    \widehat{M}_\text{b1} = E_\text{b1}(\boldsymbol{X}),
    \label{eq_baseline1}
\end{equation}
\begin{equation}
    \widehat{M}_\text{b2}  = FC_\text{b2}(E_\text{b2}(\boldsymbol{X}), \boldsymbol{q}).
\end{equation}

\subsubsection{Generalization validation}
\begin{figure}[t]
  \centering 
  \includegraphics[width=3.48 in]{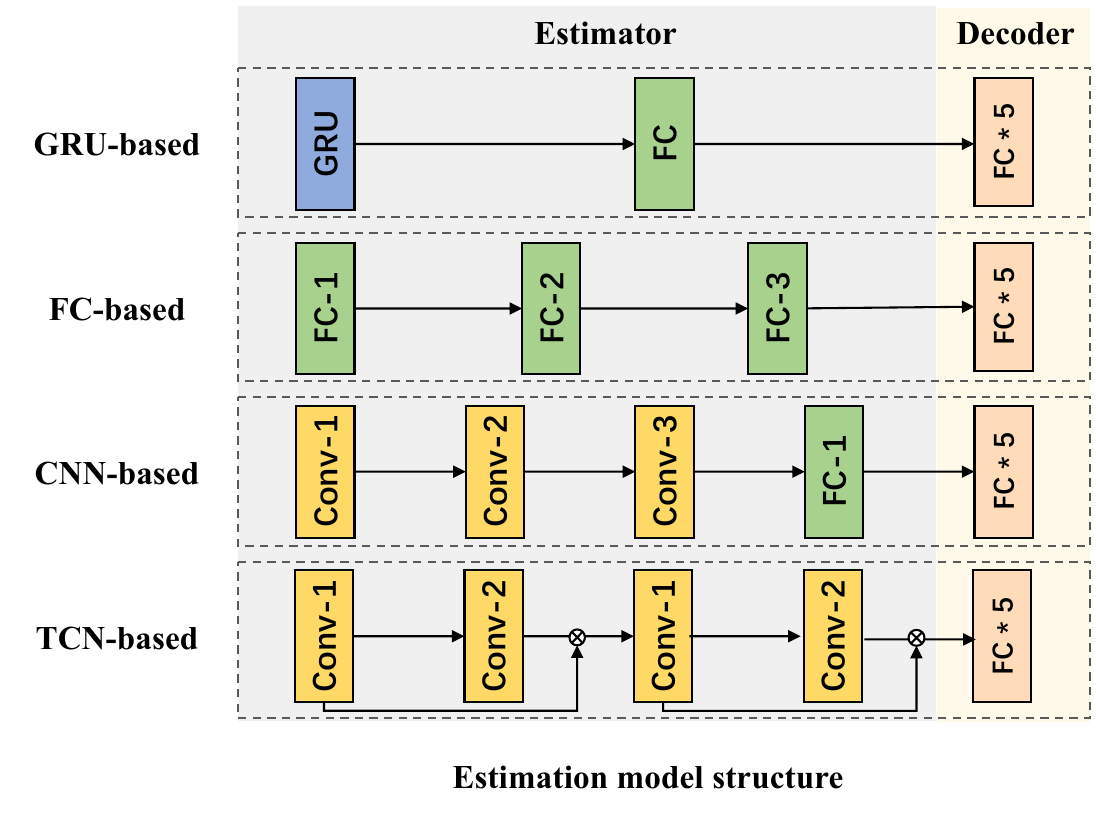}  
  \caption{Four joint moment estimation models consisting of estimators with different neural network structures, including the proposed GRU-based method for fast estimation and three mainstream model architectures.}
  \label{EstimationModelStructure}
\end{figure}

More neural network structures were used to verify the proposed GMF framework. To ensure the generalization of the proposed GMF, three existing mainstream network structures (FCNN, CNN, TCN) were further used to observe the changes in joint moment estimation accuracy after introducing GMF as shown in Fig.~\ref{EstimationModelStructure}, following the experimental process outlined in 3.2.2. 
\begin{itemize}
  \item \textbf{The FCNN} - was composed of three fully connected layers. The first layer contained 32 nodes, followed by two hidden layers with 16 nodes each.
  \item \textbf{The CNN} - consisted of three convolutional layers, each with a 3×1 kernel size and padding of 1. The layers progressively processed features with 32, 64, and 32 channels, respectively. The extracted features were then passed through an adaptive average pooling layer with an output size of 1, followed by a fully connected layer with 32 and 1 neurons.
  \item \textbf{The TCN} - employed two residual blocks, each containing two convolutional layers. The first convolutional layer within each block had a 3×1 kernel size, 32 channels, padding of 1, and a dilation of 1. The second convolutional layer maintained the same kernel size and number of channels but applied a dilation of 2 with padding of 2.
\end{itemize}

To ensure that the variables are consistent, all these models utilized the same GMF generator and decoder, which are trained with the GRU-based estimator in 3.2.2. Specifically, we replaced the GRU-based estimator with the three different network structures. Following the training process shown in Fig. \ref{overview}, a series of new training processes were conducted. During the training process, only the parameters of the new estimator were updated while the parameters of the generator and decoder were fixed (keeping the generated GMF unchanged).

Previous studies such as \cite{hossainEstimationLowerExtremity2023, molinaroAnticipationDelayedEstimation2023, zhang2023predict} have demonstrated that increasing the number of network parameters has a positive impact on the estimation accuracy of joint moment estimation. Therefore, to ensure comparability of the experimental results, we adjusted the network parameter number of the three models to make them as close as possible to the GRU-based model in terms of validation accuracy when conducting the first baseline as (\ref{eq_baseline1}). 

\subsection{Assistance for Human Walking}

The poor generalized model might lead to the controller instability, which reduces the exoskeleton performance and even results in a fall for the user \cite{molinaroEstimatingHumanJoint2024}. Therefore, real-world validation is critical for instantaneous joint moment estimation methods used for exoskeleton control. We validated the proposed method on an exoskeleton and implemented walking experiments with users wearing it.

Considering the validation effect, we adopted a rigid exoskeleton structure, ensuring the full torque transfer from the motors to the user as possible. The candidate coefficients of (\ref{eq_assist_tau}) were set as $k = 0.15$ and $\delta = 0.3$ s, which were determined through preliminary experiments. 

To quantify the gain for reducing efforts when users walk with the proposed high-level controller, comparison experiments were conducted. Five participants (age 23.4 ± 1.4 y, height 1.7 ± 0.2 m, mass 60.6 ± 1.4 kg) were recruited to complete metabolic trials under three conditions: No Exo, Assist off, and Assist On. The K5 portable metabolic system (COSMED, Rome, Italy) was used to record the experiment data. Specifically, A subject took part in 3 trials with their metabolic cost being collected: a trial without the hip exoskeleton including neutral standing for 3 minutes and walking at 1.25 m/s for 5 minutes, a trial with the no-powered exoskeleton for 5 minutes walking, and a trial with the powered exoskeleton for 5 minutes walking. At least a 10-minute break between trials to allow subjects to return to normal metabolic levels. Before each trial, the participant was required to fast for a minimum of 2 h to reduce metabolic bias from food intake. 
The Instantaneous metabolic cost $E$ was computed based on $\dot{V}O_2$ and $\dot{V}CO_2$ (collected by K5) as:
\begin{equation}
    E = \frac{16.48 * \dot{V}O_2 + 4.48 * \dot{V}CO_2}{ 60 * m},
\end{equation}
where $m$ is the body mass of the subject, and the net metabolism rate $E_\text{net}$ was obtained by:
\begin{equation}
    E_\text{net} = E_\text{walking} - E_\text{static}.
\end{equation}

\section{Results}

In this section, the results of the experiments are presented with detailed statistical analysis.
The convergence of the proposed model during training is illustrated in Subsection 4.1. Subsections 4.2 and 4.3 demonstrate the accuracy improvement brought by the proposed GMF method and the generalization performance of GMF. Subsection 4.4 shows the computation efficiency advantage of using a GRU-based estimator on the exoskeleton controller. In Subsection 4.5, the tested assistance gain for users with the proposed method is given.

\begin{figure*}[h]
  \centering 
  \includegraphics[width=\linewidth]{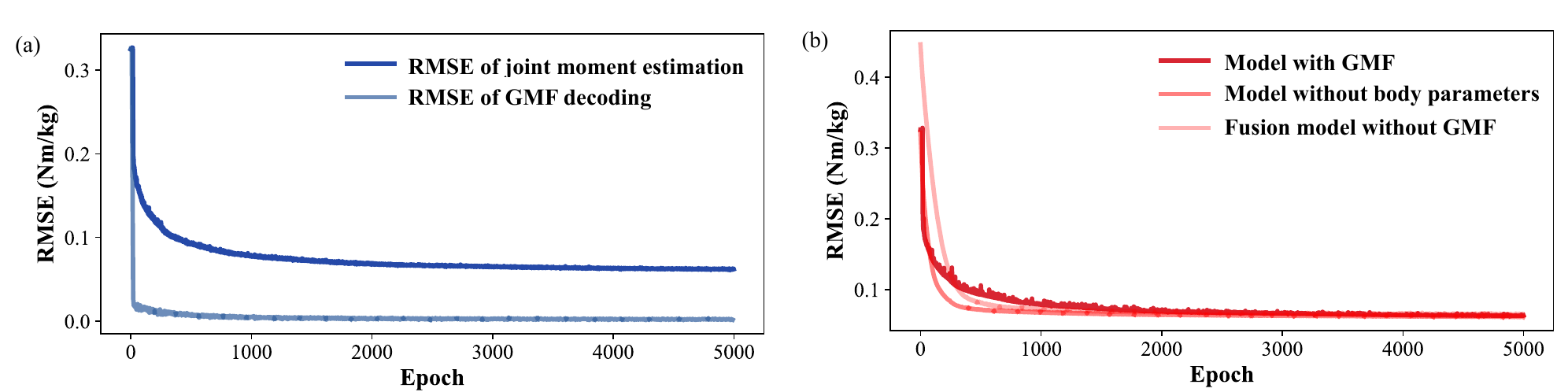}  
  \caption{(a) Trend of the accuracy of the joint moment estimation (dark blue) and accuracy of GMF decoding (light blue) of training data during the training process of the proposed estimation model based on GMF. Both values of them continuously improve and converge to the optimum over the 5000 epochs. (b) The trend of estimation accuracy on the validation set for our method (dark red) compared to two other contrastive methods (light red). The three methods approximate the best accuracy achieved under the experimental conditions set.}
  \label{loss}
\end{figure*}
\begin{figure*}[h]
  \centering 
  \includegraphics[width=\linewidth]{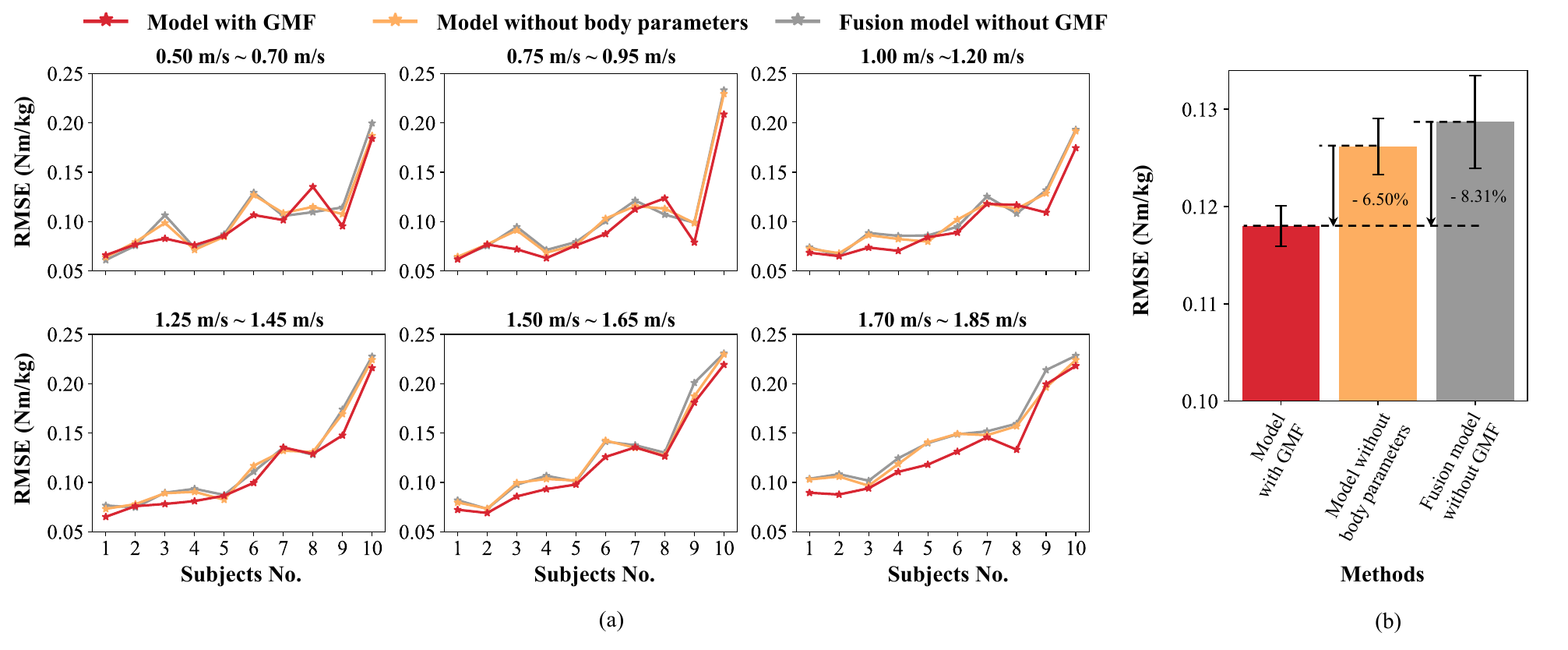}  
  \caption{(a) Comparison of the accuracy of joint moment estimation for all subjects at different speed ranges using the proposed method, along with the baseline methods. The proposed method improves the estimation accuracy of the model in most cases, especially at faster speeds. (b) Comparison of the overall average and variance of the test results for the three methods.}
  \label{rmse_result}
\end{figure*}
\subsection{Convergence of Neural Network}

The convergence of the proposed neural network trained with 10 subjects from dataset is illustrated in Fig.~\ref{loss} (a), represented by the root mean square error (RMSE):
\begin{equation}
  RMSE = \sqrt{\frac{1}{n}\sum_{i=1}^{n}(y_i - \hat{y_i})^2},
\end{equation}
\noindent where $\hat{y}_i$ denotes the estimated value by the neural network, $y_i$ is the ground truth value, and $n$ is the number of samples. During the training process, both the joint moment estimation accuracy and the decoding accuracy of the decoder consistently improved without significant fluctuations. The RMSE of the joint moment estimation on the training dataset, averaged over five repetitions, was 0.0630 ± 0.0014 Nm/kg, while the RMSE of GMF decoding was 0.0016 ± 0.0017 Nm/kg.

\subsection{Estimation Accuracy of Models}
Fig. \ref{rmse_result} reports the RMSE results of our method compared to the baseline methods. Overall, our method achieves an RMSE of 0.1180 ± 0.0021 Nm/kg on the test dataset. The mean results show a significant improvement compared to the two baseline methods (6.50\% vs. model without body parameters and 8.31\% vs. fusion model without GMF). The experimental results demonstrate that, except for subject 6, whose average joint moment estimation error of all walking speed increased by 0.97\% compared to model without body parameters and 2.57\% compared to fusion model without GMF,
the joint moment estimations for other subjects are more accurate. 
The second baseline method, which directly fuses body parameters ($\boldsymbol{q}$) and joint kinematics ($\boldsymbol{X}$) through the network, does not achieve improved joint moment estimation accuracy compared to model without body parameters with only inputting joint kinematics. In fact, its performance slightly degrades, suggesting that conventional fusion models struggle to effectively integrate body parameters.

As it is difficult to achieve the optimal validation set accuracy with limited time, in addition to using the RMSE, we used the rate of increase in the RMSE on the test set relative to the RMSE on the validation set, denoted as $R_{tv}$, which represents the degree of overfitting~\cite{hawkins2004problem,montesinos2022overfitting} of the optimal model, obtained from:
\begin{equation}
R_{tv} = \frac{RMSE_{\text{test}}-RMSE_{\text{validation}}}{RMSE_{\text{validation}}}.
\end{equation}
This equation implies that the $R_{tv}$ of a model with better generalization performance should be smaller(with consistent accuracy on the validation dataset).

Fig.\ref{loss} (b) shows the variation of validation accuracy during the training process for our method and the two baseline methods. All three models achieve similar accuracy on the validation set (around 0.0632 Nm/kg). Based on the statistical analysis (Table.~\ref{TVAR}), our method reduces $R_{tv}$ by 20.06\% and 19.20\%. While RMSE intuitively demonstrates the effectiveness of our method, the $R_{tv}$ for result evaluation reveals the underlying reason for the accuracy improvement of our method: the mapping of GMF and joint kinematics is more accurate, reducing the estimation inaccuracy of the model in new subject environments.
\begin{table}[t]
  \centering
  \caption{$R_{tv}$ of proposed method compared to baseline methods}
  \resizebox{\linewidth}{!}{
    \begin{tabular}{cccc}
    \toprule 
    & Model with GMF & \makecell{Model without \\  body parameters} & \makecell{Fusion model \\  without GMF} \\
    \midrule
    $R_{tv}$ & 0.8907 & 1.1142 & 1.1023 \\
    Rate of Increase & - & 20.06\% & 19.20\% \\
    \bottomrule 
  \end{tabular}
  }
\label{TVAR}
\end{table}

\begin{figure*}[!t]
  \centering 
  \includegraphics[width=\linewidth]{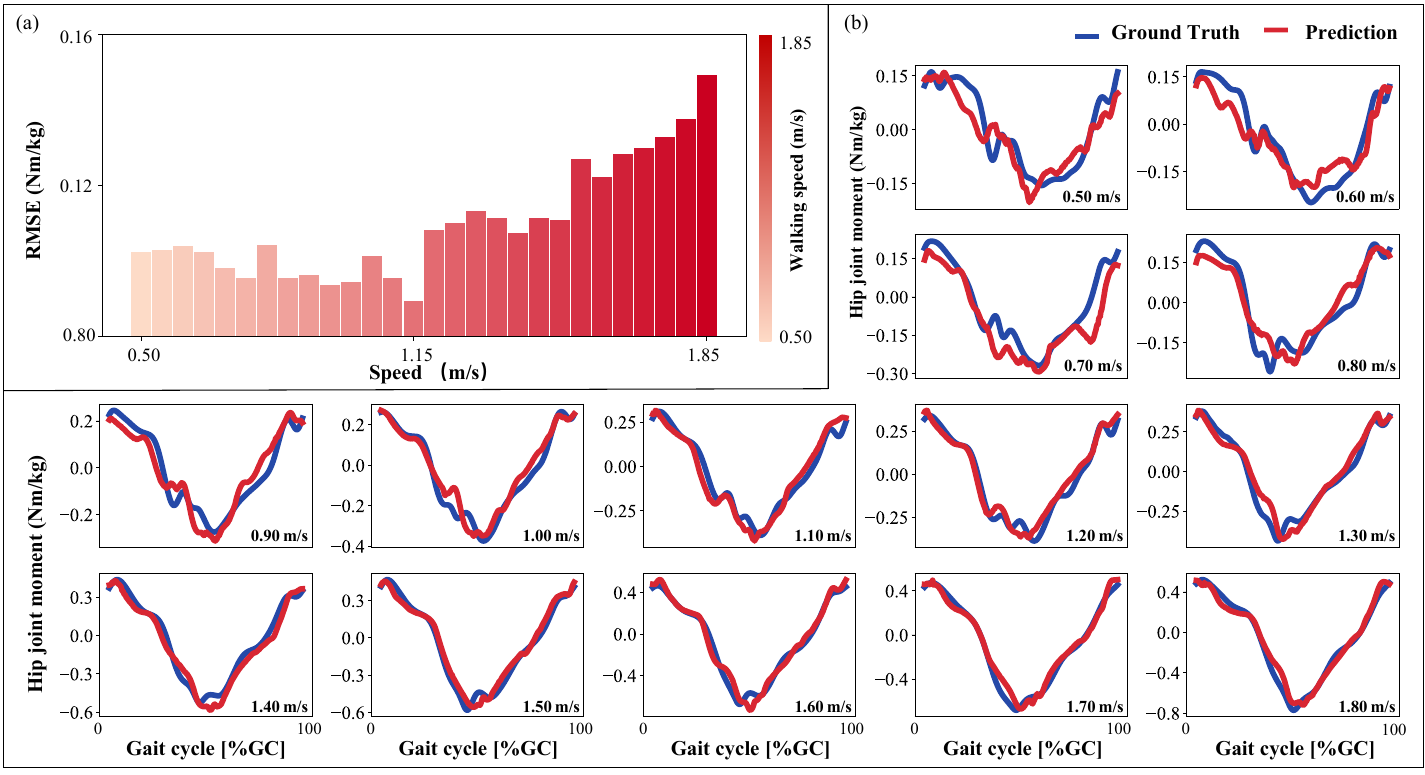}  
  \caption{(a) Joint moment estimation error of the proposed method at 28 different walking speeds. The results show that within the provided range of walking speeds, the joint moment estimation error exhibits a "V" shaped pattern, with the minimum estimation error occurring at 1.15 m/s. (b) Samples of the predicted curve and the ground truth curve of a complete gait cycle for one subject at 14 walking speeds, spaced at intervals of 0.1 m/s.}
  \label{28speed}
\end{figure*}
\begin{table*}[!h]
  \centering
  \caption{Estimation accuracy performance with different estimators in the generalization experiment}
  \begin{tabular}{cccc}
    \toprule
    Estimator & Model without body parameters & Fusion model without GMF & GMF-based model (Ours) \\
    \hline
    GRU & 0.1262±0.0021 & 0.1287±0.0048 & \textbf{0.1180±0.0030} \\
    \hline
    FCNN & 0.1270±0.0029 & 0.1324±0.0048 & \textbf{0.1169±0.0018} \\
    \hline
    CNN & 0.1231±0.0009 & 0.1270±0.0122 & \textbf{0.1131±0.0009} \\
    \hline
    TCN & 0.1260±0.0011 & 0.1270±0.0037 & \textbf{0.1154±0.0008} \\
    \hline
    Average & 0.1256±0.0015 & 0.1288±0.0022 & \textbf{0.1159±0.0018} \\
    \bottomrule
  \end{tabular}
  \label{generation}
\end{table*}

Complete hip joint moment curves for one gait cycle of a subject at 14 different walking speeds is shown in Fig. \ref{28speed} (b) for qualitative comparison between the ground truth and predicted values. Furthermore, we analyzed the variation of model estimation error across 28 walking speeds ranging from 0.5 m/s to 1.85 m/s, as depicted in Fig. \ref{28speed} (a). The estimation error exhibits a decreasing and then increasing trend. For low walking speeds, this could be attributed to the greater variability in gait patterns during slow walking \cite{kang2008separating}. On the other hand, the increase in estimation error for fast walking speeds is likely driven by the higher magnitude of joint moments, as shown in Fig.~\ref{28speed} (b). 

\subsection{Generalization of GMF}
Table. \ref{generation} presents the performance of GMF in the generalization experiment. Within the scope of this experiment, all four estimators showed an improvement in joint moment estimation accuracy after introducing GMF (compared to the method without body parameters: 7.96\% in FCNN, 8.12\% in CNN, 8.41\% in TCN, and compared to the method fusion body parameters but without GMF:  11.44\% in FCNN, 10.94\% in CNN, 9.13\% in TCN). Since the GMF generator and decoder used in the experiment were from a pre-trained model and with no parameters change, the results further demonstrate that the generated GMF is a more individual-independent representation of joint moment and contributes to the accuracy of predicting joint moment based on joint kinematics.

\begin{figure*}[t]
  \centering 
  \includegraphics[width=\linewidth]{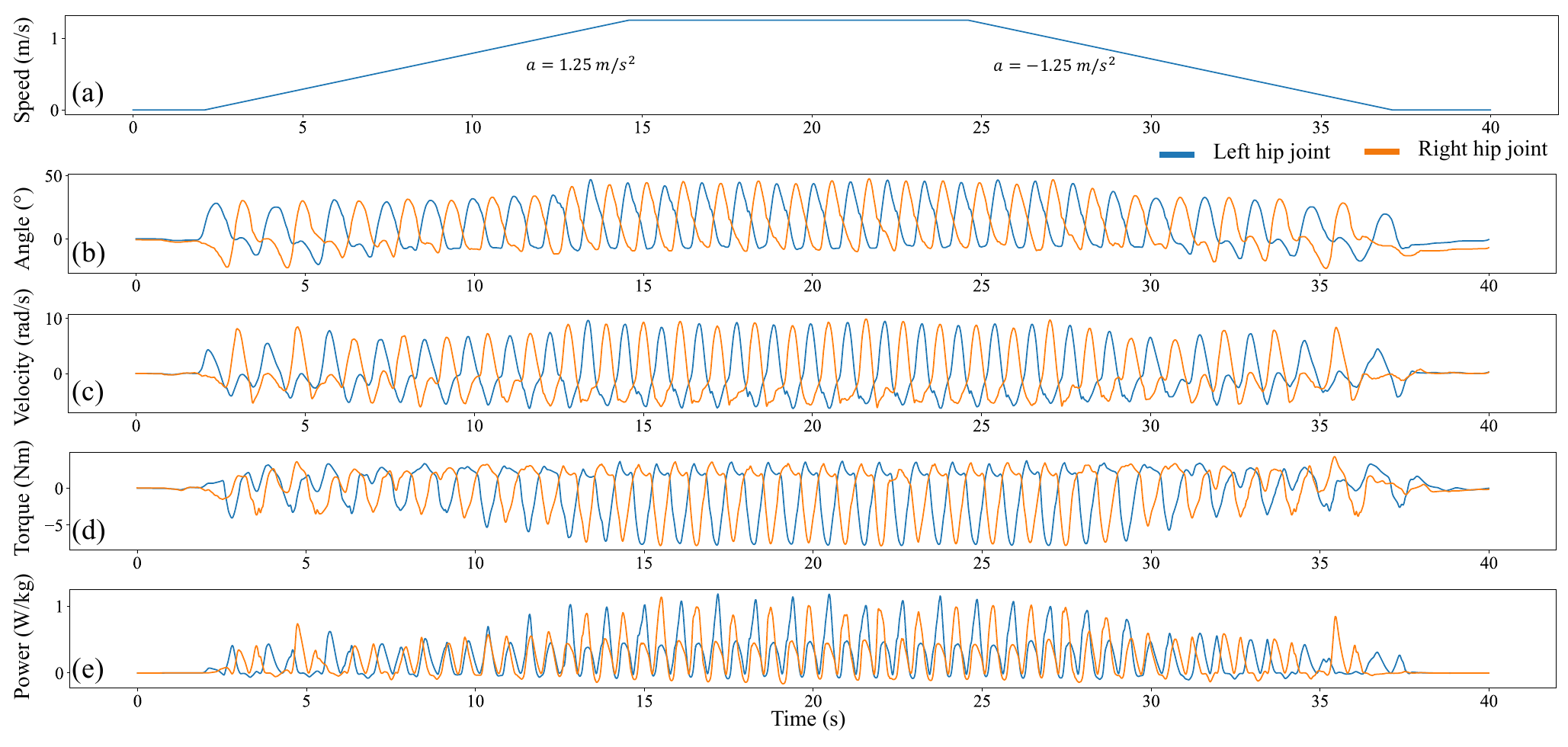}  
  \caption{Controller performance in variable-speed walking trial. (a) Walking speed varied from 0 m/s to 1.25 m/s, maintained for 10 s, and then decelerated to 0 m/s.(b)-(e) Motor angle, angular velocity, assistance torque, and power trajectories on hip joint with the walking speeds corresponding to (a). The positive direction of the vertical axis represents the flexion of the leg during walking, while the negative direction represents the extension.}
  \label{motor_property}
\end{figure*}
\begin{figure*}[!t]
  \centering 
  \includegraphics[width=\linewidth]{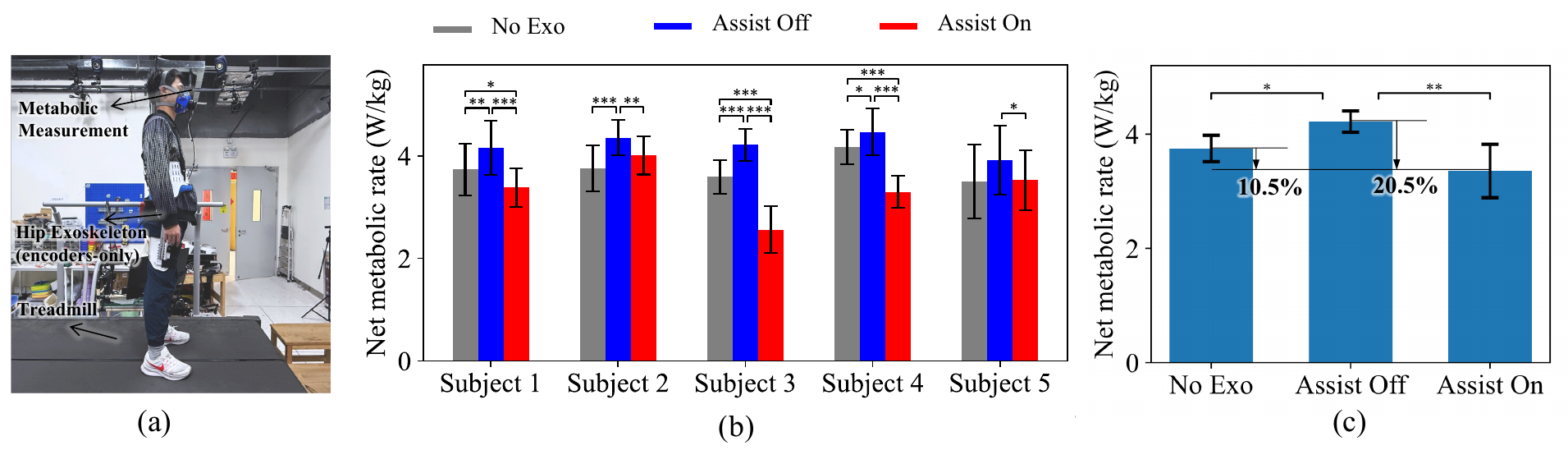}  
  \caption{(a) Experiment environment of the metabolic cost test. (b) Net metabolic rate for five subjects (*: p$<$0.05, **: p$<$0.01, ***: p$<$0.001). All subjects received the assistance gain compared to Assist Off. (c) Average net metabolic rate for all subjects}
  \label{metabolismExp}
\end{figure*}
\subsection{Reducing Metabolism Rate of Users during Walking}
Fig. \ref{motor_property} illustrates the assistance performance of the exoskeleton deployed with proposed GMF-based controller, including the assistive torques during three phases: transitioning from standing to walking, steady-state walking at 1.25 m/s, and transitioning from walking to standing. Benefiting from the powerful representational capabilities of neural networks, the GMF model controller can achieve stable assistance during process transitions. The peak torque was about 7.81 Nm for the support phase and 4.29 Nm for the swing phase, respectively. The peak amplitude Difference between the two phases shows the proposed method achieves expected assistance torque as it is widely known the hip joint requires greater moment in the support phase than it in the swing phase during walking. Fig.~\ref{motor_property} also shows a more detailed manifestation of the hip joint angle, angular velocity, and power during assistive walking. All of these data were obtained from the motor encoders of hip exoskeleton. 

The metabolic cost test results are presented in Fig.~\ref{metabolismExp}.
During the experiments, the exoskeleton, utilizing the proposed joint moment estimation method, delivers assistive torque to users, effectively reducing their metabolic rate. The mean net metabolic cost reduction achieves 10.5\% compared to normal walking (No Exo), while the reduction achieves 20.5\% compared to the Assist Off model ($p<0.01$). A notable result is that 2 participants (Subject 3 and Subject 4) in the validation experiments achieved better assistance gain because the participants were most comfortable with the exoskeleton size and had undergone longer walking training for using the exoskeleton, which has been proven to be effective in increasing a subject's metabolic cost reduction \cite{poggensee2021adaptation}.

\section{Discussion}
This study utilizes a GMF generation method with a GRU-based network to achieve a better estimation of the hip joint moment and its application on the exoskeleton, which benefits from the consideration of inter-user variability in joint moment estimation for exoskeleton assistance.
The GMF, which is obtained through an adaptive neural network, leads to significant improvements in model accuracy and has the potential to become an advanced method for future joint moment estimation. The metabolism cost test result showed that our methods achieve the satisfactory results of assistance gain.

\subsection{Balance Between Accuracy and Computation Efficiency}
This paper proposed a GMF-based neural network framework, aiming to improve estimation accuracy with limited computation sources. To achieve the accuracy superior to the state-of-the-art (SOTA) model, neural networks with large-scale parameters are conventionally required. Our previous research has stated the relationship between accuracy and computation time \cite{zhang2023predict}, in which the estimation accuracy of the neural network approaches saturation as the number of network parameters increases. For a real-time exoskeleton control system, efficient computation translates to faster response speed, higher estimation precision, and longer endurance time. This is one of the primary motivations of this study: to explore more efficient estimation methods rather than relying solely on increasing the number of neural network parameters.

It is clear that better estimation accuracy would benefit the exoskeleton control based on instantaneous joint moment estimates. However, there is an uncertain question of whether a better prediction accuracy provides a significant increase in assistance gain.
From the results in Table.~\ref{generation}, FCNN, CNN and TCN models have a higher estimation accuracy than the GRU model. However, on the one hand, the other three models have a lower control frequency, which may lead to a poor performance when they were applied on the hip exoskeleton. On the other hand, the multiple comparisons test (with Bonferroni correction) results show that the difference in estimation accuracy among the four models is not completely significant. Therefore, the GRU model is selected as the GMF estimator for the exoskeleton control system in this study.

\subsection{Transferability of Using Other Input Sensors}
Compared to most of the existing methods for estimating joint moment (using sensor information such as IMU as model input), the GRU-based estimator used in this paper takes joint kinematics data as inputs instead of raw sensor data. Because the joint moment estimation had been considered a model identification problem that requires establishing an approximate model with dynamic characteristics rather than a completely uninterpretable black-box model. This also means that the model controller in this study uses sensor data in a more humanly understandable way, trying to ensure the reliability of the result \cite{kang2022reduce}. Our exoskeleton experiments were able to implement a deep learning-based assistance control model utilizing only motor encoders and got successful results, which we believe it was benefited from this more rational approach to data processing. 

Besides, for other potential practical applications, model training can be performed using angles obtained from sensors or using highly correlated data like IMU data of sensors as inputs directly, without requiring changes to the overall model structure. However, additional challenges may be posed in these ways, such as sensor accuracy drift and data transmission loss \cite{zhu2024real}.

 
\subsection{Better Potential Estimation Accuracy}
The other important inputs used by the model are human body parameters, which are composed of height and weight in this paper. These two parameters are valid but might not be the optimal attributes for obtaining GMF. Future research could explore more suitable body parameters to achieve better GMF generation and more accurate estimations in the problem of joint moment estimation based on joint kinematics.

Furthermore, since the proposed method originates from the research on assistive hip exoskeleton control, the estimator model proposed in this paper uses a GRU-based network structure. However, for joint moment estimation tasks that do not require timeliness, the optimal network structure for the estimator may not necessarily be based on recurrent neural networks. The joint moment estimation method we propose can be extended to the field of biomechanics, as demonstrated in the generalization experiment, where it also performs well on mainstream neural networks besides the GRU-based network. For systems with low real-time requirements, future researchers can use more appropriate network structures within the framework of this method to obtain higher-precision joint moment estimation models.

\subsection{Limitations}
The proposed method still has several limitations that need to be further addressed in future research.
\subsubsection{Burden of Wearing Exoskeleton}
Fig.~\ref{metabolismExp} has shown a 12.5\% metabolic cost increase when subjects wearing the hip exoskeleton without power compared to normal walking. Although this study focused on achieving accurate joint moment estimation, there is a clear need for exoskeleton structures with low wearing burden to enhance the assistance gain for users. Meanwhile, as the results of the experiment more conclusively verified that training benefits the exoskeleton user to reach a higher assistance gain, future studies may need to consider how to control this impact in experiments.
\subsubsection{Accuracy of the Decoder}
To decode the GMF, we use a fully connected neural network with the same structure as the feature generator, and make the loss of decoder back-propagate to the generator with a weighting coefficient to ensure the decodability of the generated GMF. While this method achieves high accuracy (decoding RMSE less than 0.003 Nm/kg) in decoding the GMF into the joint moment, the decoding error can only approximate zero rather than reaching it due to the use of a separate neural network. Additionally, to ensure accurate GMF decoding, the decoder may become overly complex, leading to increased computational time. This complexity transforms what should be a straightforward inverse transformation into a more intricate and less reliable process. The concept of reversible neural networks \cite{jacobsenIRevNetDeepInvertible2018, gomezReversibleResidualNetwork, dinhNICENonlinearIndependent2015a}, which have lightweight memory consumption and can preserve input information features, has been proposed. When applied to our method, it has the potential to achieve zero error in GMF decoding with fewer parameters. We may further explore the application of such networks in our method in the future.

\subsubsection{Challanges of Exoskeleton Control with Encoder-only Sensors}
Inspired by the idea that any sensor capable of measuring joint kinematics can be used to estimate joint moments, we implemented exoskeleton control using motor encoders alone. However, encoder-only sensors still face significant limitations in exoskeleton applications, primarily due to the known issue of coupling between motor motion and human movement.

Our experiments show that encoder-only sensors can provide stable and accurate estimations of joint moments when the assistance time is appropriately calibrated and comfortable for the user. However, determining the correct assistance timing requires pre-experimentation to identify the optimal delay time factor. Furthermore, if the assistance time is not accurately adjusted, the assistive ratio can significantly amplify the negative effects, leading to suboptimal performance.
To address this issue in the future, more specific methods should be developed, such as the use of Series Elastic Actuators (SEA)~\cite{qianAdaptiveOscillatorBasedAssistive2022}, abnormal assistance detection techniques, and other potential solutions.

\section{Conclusion}
To address the issues of lack of personalization joint moment estimation neural network for hip exoskeleton assistance, this study proposes an estimation method based on GMF generation. The method utilizes a generator to adaptively generate GMF, which serves as a more individual-independent representation compared to the biomechanical joint moment. A GRU-based neural network is then employed to estimate the generated GMF. The GMF improves the estimation accuracy of the estimator while could be decoded into joint moments accurately.
Experimental results demonstrate enhanced estimation accuracy and strong generalization capability of the proposed method. Furthermore, real-world exoskeleton tests using encoder-only sensors validate its effectiveness in delivering assistive benefits to users. The contributions of this work advance the feasibility of real-time joint moment estimation and its practical application in exoskeleton control.
\section*{Acknowledgments}
We sincerely thank all the students in HARLab for their invaluable support of this work, as well as all the participants in our experiments for their time and effort.
\subsection*{Author Contributions} 
Yuquan Leng and Chenlong Fu supervised the project.
Yuanwen Zhang and Jingfeng Xiong conceived the idea and designed the experiments.
Chuheng Chen and Xinxing Chen reviewed the manuscript and proposed refinements of the experiment.
All authors reviewed the manuscript and approved the version to be submitted.

\subsection*{Funding}

This work was supported by National Key R\&D Program of China [grant numbers 2024YFC3082800]; the National Natural Science Foundation of China [grant numbers 52175272]; Guangdong Basic and Applied Basic Research Foundation [grant numbers 2024B1515020008 and 2023B1515130007]; and Shenzhen Science and Technology Program [grant numbers KCXFZ20230731093401004, RCYX20231211090345058 and JCYJ20220530114809021].
\subsection*{Declaration of interest}
None
\subsection*{Data Availability}
All data needed to evaluate the conclusions are presented in the manuscript and/or the Supplementary Materials. Additional data that support the findings of this study are available from the corresponding author upon reasonable request

\section*{Supplementary Materials}

Movies S1 and S2.
\bibliographystyle{elsarticle-num} 
\bibliography{citeBB.bib}

\end{document}